\documentclass{llncs}

\usepackage[utf8]{inputenc}
\usepackage{url,hyperref}
\usepackage{graphicx}

\usepackage[draft]{fixme}
\fxsetup{theme=color, layout={inline,index}}
\usepackage{booktabs}
\usepackage{amsmath}

\usepackage{color}
\usepackage{amsmath}
\usepackage{amssymb}
\usepackage{bm}
\usepackage{siunitx}
\usepackage{tabularx}
\usepackage{tabu}
\usepackage{graphicx}
\usepackage{subfig}
\usepackage{threeparttable}

\DeclareMathOperator*{\argmin}{arg\,min}

\newcommand{\reffig}[1]{Fig.~\ref{#1}}

\usepackage[normalem]{ulem}
\useunder{\uline}{\ul}{}

\usepackage{tikz}
\usetikzlibrary{arrows}
\usetikzlibrary{positioning,calc}
\usetikzlibrary{decorations.pathreplacing}
\usetikzlibrary{decorations.markings}
\usetikzlibrary{fit}
\usetikzlibrary{shapes.callouts}
\usetikzlibrary{shapes.geometric}
\usetikzlibrary{matrix}

\tikzset{
  big arrow/.style={
    decoration={markings,mark=at position 1 with {\arrow[scale=1.4]{latex}}},
    postaction={decorate},
    shorten >=2.5pt}}
\tikzset{
  big 2arrow/.style={
    decoration={markings,mark=at position 0 with {\arrow[scale=-1.4]{latex}},mark=at position 1 with {\arrow[scale=1.4]{latex}}},
    postaction={decorate},
    shorten >=2.5pt}}
    
\pgfdeclarelayer{background}
\pgfdeclarelayer{foreground}
\pgfsetlayers{background,main,foreground}

\usepackage{tablefootnote}

\begin{document}

\frontmatter          % for the preliminaries

\pagestyle{headings}  % switches on printing of running heads
\addtocmark{KittingBot – Robust Perception and Manipulation Planning for Part Kitting in Automotive Logistics} % additional mark in the TOC

\mainmatter              % start of the contributions

\title{KittingBot: A Mobile Manipulation Robot for Collaborative Kitting in Automotive Logistics}
\titlerunning{KittingBot: Collaborative Kitting in Automotive Logistics}  % abbreviated title (for running head)
\author{Dmytro Pavlichenko \and Germ\'an Mart\'in Garc\'ia \and Seongyong Koo \and \\ Sven Behnke}
\authorrunning{D. Pavlichenko, G. Mart\'in, S. Koo, and S. Behnke} % abbreviated author list (for running head)
% \tocauthor{Ivar Ekeland, Roger Temam, Jeffrey Dean, David Grove,
% Craig Chambers, Kim B. Bruce, and Elisa Bertino}
%
\institute{Autonomous Intelligent Systems, Computer Science Institute VI, University of Bonn, Endenicher Allee 19a, 53115 Bonn, Germany\\
\email{pavlichenko@ais.uni-bonn.de}}

\maketitle              % typeset the title of the contribution

% Leave a blank line between paragraphs instead of using \\

%%%%%%%%%%%%%%%%%%%%%%%%%%%%%%%%%%%%%%%%%%%%%%%%%%%%%%%%%%%%%%%%%%%%%%%%%%%%%%%%%%%%%%%%%%%%%%%%%%%%%%%%%%%%%%%%%%%%%%%%%%%%%%%%%%%%%%%%%%%%%%%%%%%%%%%%%%%%%%%%%%%%%%%%%%%%%%%%%%%%%%%%%%%%%%%%%%%%%%%%%%%%%%%%%%%%%%%%%%%%%%%%%%%%%%%
%%% The sections below are for reference only.
%%%
%%% For Original Research Articles, Clinical Trial Articles, and Technology Reports the section headings should be those appropriate for your field and the research itself. It is recommended to organize your manuscript in the
%%% following sections or their equivalents for your field:
%%% Abstract, Introduction, Material and Methods, Results, and Discussion.
%%% Please note that the Material and Methods section can be placed in any of the following ways: before Results, before Discussion or after Discussion.
%%%
%%%For information about Clinical Trial Registration, please go to http://www.frontiersin.org/about/AuthorGuidelines#ClinicalTrialRegistration
%%%
%%% For Clinical Case Studies the following sections are mandatory: Abstract, Introduction, Background, Discussion, and Concluding Remarks.
%%%
%%% For all other article types there are no mandatory sections.
%%%%%%%%%%%%%%%%%%%%%%%%%%%%%%%%%%%%%%%%%%%%%%%%%%%%%%%%%%%%%%%%%%%%%%%%%%%%%%%%%%%%%%%%%%%%%%%%%%%%%%%%%%%%%%%%%%%%%%%%%%%%%%%%%%%%%%%%%%%%%%%%%%%%%%%%%%%%%%%%%%%%%%%%%%%%%%%%%%%%%%%%%%%%%%%%%%%%%%%%%%%%%%%%%%%%%%%%%%%%%%%%%%%%%%%

\begin{abstract}

Individualized manufacturing of cars requires kitting: the collection of individual sets of part variants for each car. This challenging logistic task is frequently performed manually by warehouseman.
We propose a mobile manipulation robotic system for autonomous kitting, building on the Kuka Miiwa platform which consists of an omnidirectional base, a 7\,DoF collaborative iiwa manipulator, cameras, and distance sensors. 
Software modules for detection and pose estimation of transport boxes, part segmentation in these containers,
recognition of part variants, grasp generation, and arm trajectory optimization have been developed and integrated.
Our system is designed for collaborative kitting, i.e. some parts are collected by warehouseman while other parts are picked by the robot. To address safe human-robot collaboration, fast arm trajectory replanning considering previously unforeseen obstacles is realized.
The developed system was evaluated in the European Robotics Challenge~2, where the Miiwa robot demonstrated autonomous kitting,
part variant recognition, and avoidance of unforeseen obstacles.

\end{abstract}

% to ease the writing process, each section should reside in its own file
\section{Introduction}
\label{sec:introduction}

Although robot manipulators and autonomous transport vehicles are widely used in manufacturing, there are still plenty of repetitive tasks, which are performed by human workers. Automation of such tasks would allow for relieving workers from repetitive and dull activities, which may cause harm to their health. Furthermore, automation has the potential to increase productivity and quality.

\begin{figure}[tb]
\centering
  a)\,\includegraphics[height=0.4\linewidth]{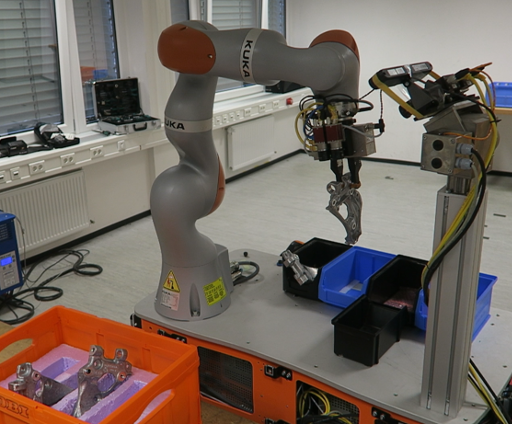}\hspace*{3ex}
  b)\,\includegraphics[height=0.4\linewidth]{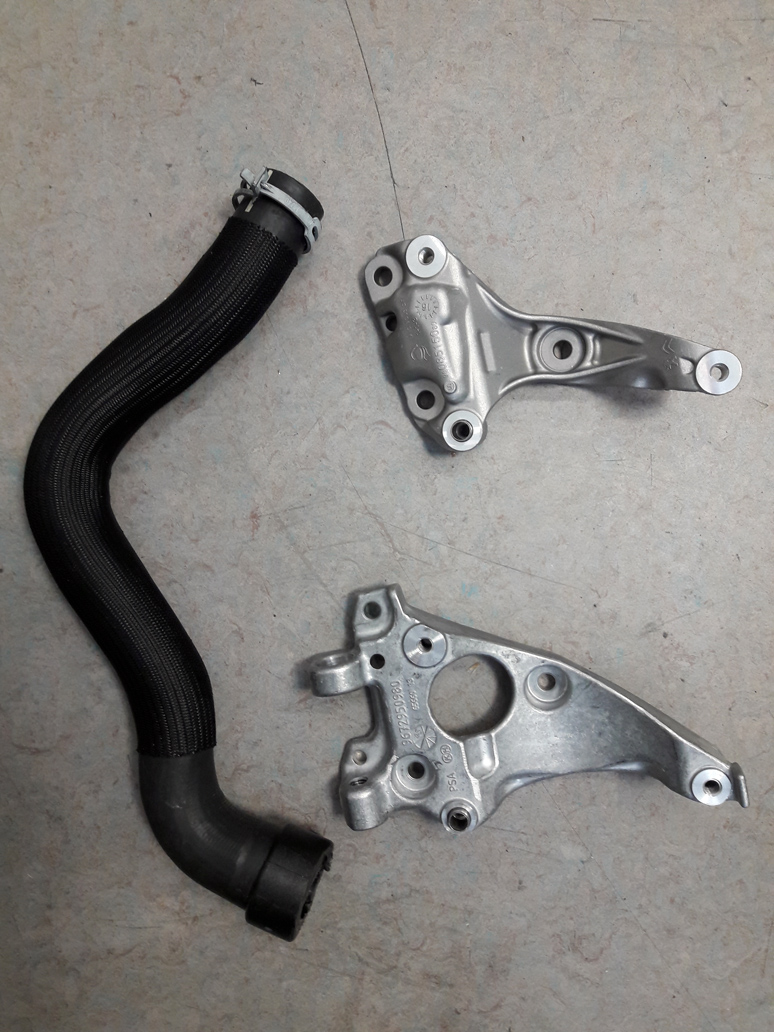} \vspace*{-2ex}
  \caption{
  a) Kuka Miiwa (KMR iiwa) robot performing the kitting; 
  b) Three types of parts used in the kitting task.}
  \label{fig:kitting_intro}
\end{figure}

In this paper, we address the task of kitting in automotive logistics, which is frequently performed manually by warehouseman. Kitting became necessary, because car manufacturing has been individualized. Each customer configures its car, such that for each car sets of part variants must be collected and delivered to the assembly line just in time. Kitting is performed in a large storage area, called automotive supermarket, where all part variants can be collected from transport boxes and pallets.
For each manufactured car, an individual order for the needed part variants is generated. A warehouseman collects the parts in the automotive supermarket and sends them to the assembly line as a kit.

We propose a mobile manipulation robotic system for autonomous kitting, building on the Kuka Miiwa platform~\cite{domel2017},  which consists of an omnidirectional base, a 7\,DoF collaborative iiwa manipulator, cameras, and distance sensors. 
The robot performing the kitting and three parts representing the kit are shown in \reffig{fig:kitting_intro}.

In order to effectively solve the kitting task, several subtasks must be addressed. First of all, the robot has to navigate precisely within the supermarket in order to reach target locations for part collection. Upon arrival at the location, it is necessary to detect the container with parts and to estimate its pose. The robot has to detect the parts inside the container and must plan suitable grasps. Collision-free arm motions must be planned and executed and the part has to be placed in the kit which is transported by the robot.

Developing an autonomous robotic system for kitting is a challenging task, because of the high degree of variability and uncertainty in each of its subtasks.
Our system is designed for collaborative kitting, i.e. some parts are collected by warehouseman while other parts are picked by the robot. To address safe human-robot collaboration, fast arm trajectory replanning considering previously unforeseen obstacles must be realized.

In this paper, we present our approaches for perception and manipulation, as well as system integration and evaluation.
 We solve the perception task with a robust pipeline, consisting of the steps:
\begin{itemize}
	\item Container detection and pose estimation,
	\item Part segmentation and grasp generation, and
	\item Classification of the grasped part before it is put into the kit---to verify that it is the correct part.
\end{itemize}

In order to perform manipulation effectively, we utilize an arm trajectory optimization method with a multicomponent cost function,
which allows for obtaining feasible arm trajectories within a short time.
All developed components were integrated in a KittingBot demonstrator. 
The developed system was evaluated at the Showcase demonstration within the European Robotics Challenge~2: Shop Floor Logistics and Manipulation\footnote{EuRoC Challenge~2: \url{http://www.euroc-project.eu/index.php?id=challenge_2}}, where we participated as a challenger team together with the end user Peugeot Citro\"{e}n Automobiles S.A. (PSA)\footnote{Peugeot Citro\"{e}n Automobiles S.A.: \url{https://www.groupe-psa.com}}. 
We report the success rates for each step of the kitting pipeline as well as the overall runtimes.

\section{Related Work}
\label{sec:related_work}

In recent years, interest in mobile manipulation robots increased significantly.
Many components necessary for building an autonomous kitting system have been developed.
Integrating these to a system capable of performing kitting autonomously is challenging, though.
In this section, we give overview of autonomous robotic systems for kitting in industrial environments.

An early example of mobile bin picking has been developed by Nieuwenhuisen et al.~\cite{Nieuwenhuisen:ICRA2013}. 
They used the cognitive service robot Cosero~\cite{Stuckler:Frontiers2016} for grasping unordered parts from a transport box and delivering them to a processing station. 
Part detection and pose estimation was based on depth measurements of a Kinect camera and the registration of graphs of geometric primitives~\cite{Berner:ICIO2013}. Grasp planning utilized local multiresolution representations. 
The authors report successful mobile bin picking demonstrations in simplified settings, but their robot was far from being strong and robust enough for industrial use.

Krueger et al.~\cite{7440782} proposed a robotic system for automotive kitting within the STAMINA\footnote{European FP7 project STAMINA: \url{http://stamina-robot.eu}} project.
The large mobile manipulation robot consists of an industrial manipulator mounted on a heavy automated guided vehicle (AGV) platform. 
The system utilizes the software control platform SkiROS~\cite{Rovida2017} for high-level control of the mission, which is composed of skills~\cite{Mikkel2016}.
Each skill solves a specific sub-task: i.e. detecting a part, generating a grasp, etc.~\cite{Holz2017}.
Such an architecture allows for fast definition of the global kitting pipeline for each specific use case, which can be performed by the end user.
Crosby et al.~\cite{Crosby2017} developed higher-level task planning within the STAMINA project.
SkiROS is used to bridge the gap between low-level robot control and high-level planning.
The STAMINA system was tested in a simplified setting within the assembly halls of a car manufacturer, where the  robot successfully performed full kitting procedures for kits of one, three, four, and five parts multiple times. Execution speed was slow, though, and safe human-robot collaboration has not been addressed.

Krug et al.~\cite{Krug2016} introduced APPLE---a system for autonomous picking and palletizing based on a motorized forklift base.
The system is equipped with a Kuka iiwa manipulator.
The authors propose a grasp representation scheme which allows for redundancy in the target gripper pose~\cite{Berenson2011}~\cite{Gienger2008}.
This redundancy is exploited by a local, prioritized kinematic controller which generates reactive manipulator motions on-the-fly.
The system has a basic safety laser scanner as well as a camera-based system for human detection.
Human workers are assumed to be wearing special reflective clothing.
Tests showed that the system is capable of performing pick and place tasks in a human-safe manner.
An interesting example of such a task was to first pick an empty pallet with a forklift, navigate to the loading zone, load the pallet using the arm, 
and finally transport the loaded pallet to the destination zone.

A similar task routine is performed in a completely different application domain: medical care.
In hospitals, nurses are required to collect necessary supplies and deliver them to the patients.
This task creates a constant dull workload for nurses, who could spend the working time in a much more patient-oriented way.
Diligent Robotics\footnote{Diligent Robotics: \url{http://diligentrobots.com}} designed a robot which should perform this routine.
The hospital environment in many cases is more challenging than industrial production lines, since the narrow corridors are often crowded with patients.

Srinivasa et al.~\cite{Srinivasa2016} address mobile manipulation tasks in household environments.
HERB---a dual-armed robot with a human-like upper-body---is used for this purpose.
The authors compose a manipulation planning module %~\cite{Bobrow1985}~\cite{Geraerts2006} 
out of several popular planners and trajectory optimizers.
This allows to effectively perform complex manipulation tasks.
For instance, the approach has been tested with a task when the robot has to load a plate, a bowl, and a glass into a tray.
Finally, the tray has to be lifted for further transportation.
The last operation required the use of both arms.
In order to configure the high-level planner, the user has to specify an action graph.

One unresolved issue with all of the above systems is that due to the large number of different objects, a large variety of grasps is required to safely manipulate them.
A robotic system with automatically exchangeable grippers does not seem to be a feasible solution, since there may be dozens of different grippers needed.
Another possible solution would be to use two grippers with the flexibility of human hands, but these are not available.
Our approach to this issue is collaborative kitting: using a simple and robust robotic gripper for picking parts with simple structure in collaboration with warehouseman who pick more complex or fragile parts.

\section{System Overview}
\label{sec:system_overview}

\begin{figure}[t]
  \centering
     \includegraphics[width=10cm]{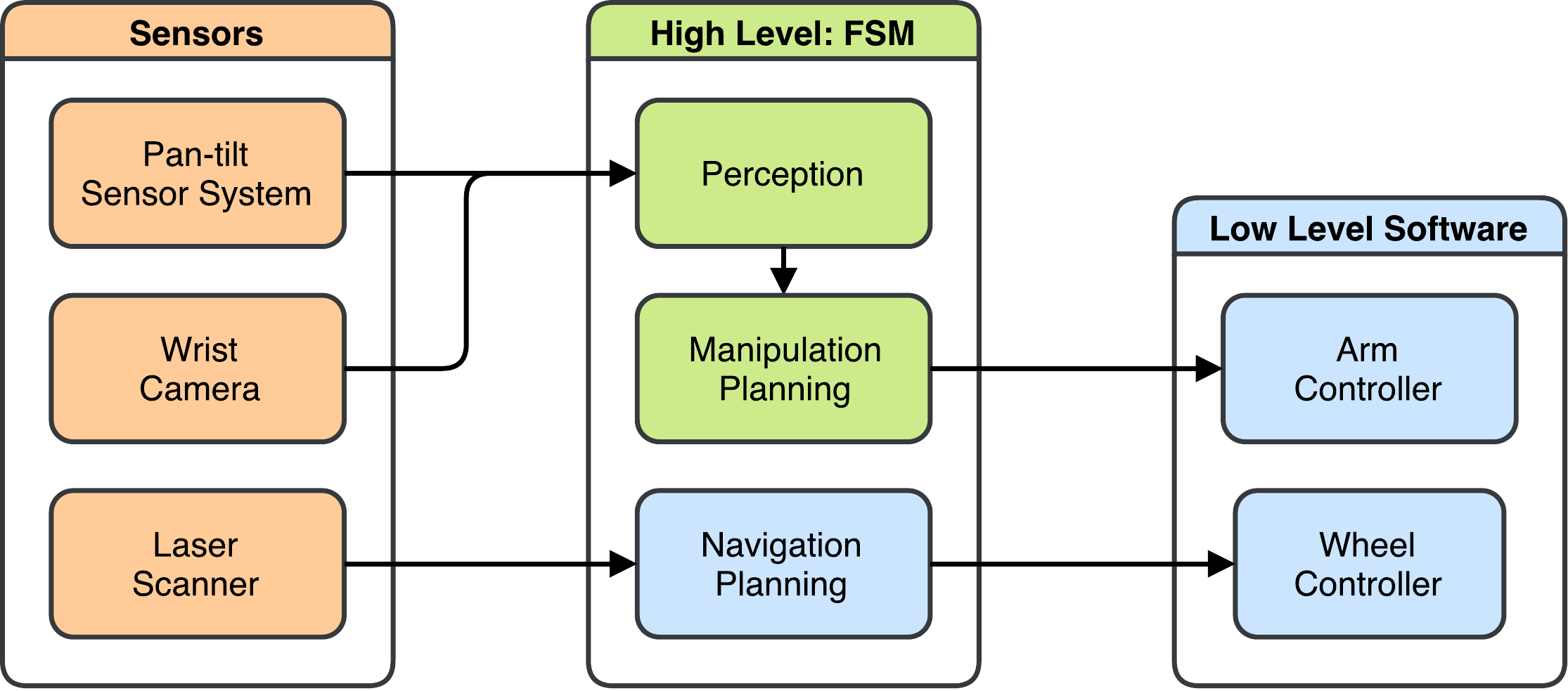}\vspace*{-1ex}
  \caption{ Simplified scheme of the proposed kitting system. Light-blue: Kuka software components. Green: components developed by us.}
  \label{fig:system}
\end{figure}

The developed system is based on the Kuka Miiwa (KMR iiwa) robot. 
The robot has a compact omnidirectional base with four Mecanum wheels.
The omnidirectional drive allows for a very precise and smooth navigation even in the areas with limited free space.
The base is equipped with multiple laser scanners on its sides in order to produce 360$^\circ$ distance measurements.
On the top surface of the base, the 7 Degrees of Freedom (DoF) collaborative Kuka iiwa arm and a vertical sensor pole with pan-tilt unit (PTU) are mounted.
The PTU carries a stereo camera system and a time-of-flight (ToF) depth camera.
The components of this system complement each other and thus avoid sensor-specific problems.
This sensor system is further referenced as Pan-Tilt Sensor System (PTSS).
A stereo camera is attached at the wrist of the iiwa arm.
While the PTSS allows to have a global view on the manipulation workspace, the wrist camera allows to measure the manipulated objects more precisely.
In order to effectively process the data from all the sensors, the robot has four Core-i7 onboard computers as well as an FPGA for the stereo processing.
The top surface of the robot base is flat and has a lot of free space, which is used to place a kit storage system.
In our experiments, we use a very simple kit storage system: three plastic boxes.

\reffig{fig:system} illustrates the main components of our system and the information flow between them.
The Kuka KMR iiwa robot comes with a low-level software stack, as well as a higher-level navigation and mapping stack.
We used these components together with ours in order to realize a complete robotic system for autonomous kitting.
The highest level of the software stack is represented by a finite state machine (FSM).
Its parameters define the whole kitting procedure: how many parts to pick up, where to pick up, where to deliver, etc.
The FSM orchesters work of all three main components of our system: perception, manipulation planning, and navigation planning.
The perception component uses sensory input from the wrist camera and the PTSS to detect the container with parts, estimate its pose, detect the parts, and define the grasp.
The manipulation planning module takes as input raw 3D sensory data for collision avoidance as well as results from the perception module.
Finally, the manipulation module produces an arm trajectory to reach the grasp and to deliver the part into the kit.
The navigation module performs mapping and path planning, as well as dynamic obstacle avoidance.

\section{Perception}
\label{sec:perception}

The location of transport boxes and pallets in the automotive supermarket is known in advance only to a limited degree of precision: boxes are manually placed and their pose can change while picking parts, placing other boxes, etc.
Hence, it is necessary to estimate the exact pose of the box in the environment.
Similarly, part poses within the containers vary and wrong part variants might be accidentally placed in the containers. 
In this section, we present the methods used for the perception of containers, segmentation of parts, and part variant recognition.

\subsection{Container Detection}
\label{sec:container_detection}

We use the approach of Holz et al.~\cite{dirk2016} for the detection and localization of containers in RGB-D data.
The method is tailored for finding containers when the upper part of the container is visible.
It is based on extracting lines along the edges in the RGB-D image and finding the best fitting models of the container we are looking for.
The container detection and localization pipeline is organized in three stages:
\begin{itemize}
	\item Detect edges in both the color image and the depth image,
	\item Fit lines to the detected edge points, and
	\item Sample subsets of lines and fit parametrized models of the containers to the subset.
\end{itemize}
The best fitting candidate gives both the lines forming the top of the box and the pose of the box. 

\subsubsection{Edge detection}

We follow the approach of Choi et al.~\cite{choi2013} for detecting edges in RGB-D data.
The method proposes the Canny edge detector for finding edges $E_{RGB}$ in the color image.
In the depth image, we inspect the local neighborhood of points, focus on points at depth discontinuities, and identify occluding edges by selecting those points $E_D$ that are closer to the camera.
In addition, we efficiently compute local covariance matrices using a method based on integral images~\cite{dirk2012}.
From the local covariance matrices, we compute local surface normals and curvature to obtain convex $E_{conv}$, and concave edges.
For the next processing steps, we combine all points at color edges, occluding edges, and convex edges to a set of edge points $E = E_{RGB} \vee E_D \vee E_{conv} , E \subseteq P$, where $P$ is a point cloud.

\subsubsection{Line detection}

Our line detection approach is based on RANSAC.
On each iteration, we select two points, $p$ and $q$, from the set $E$ and compute a line model: point on the line $p$ and direction of the line $q - p$.
We then determine all inliers in $E$ which support the line model by having distance to it below threshold $\epsilon_d$.
The line model with the largest number of inliers is selected as the detected line $l$.
If the number of inliers of line $l$ exceeds the minimum number of inliers, $l$ is added to the set of lines $L$.
We then remove the inliers of $l$ from $E$ and continue detecting further lines.
If the residual number of points in $E$ falls below a threshold, or the minimum number of inliers for the line segments is not reached, the line detection is stopped.

\subsubsection{Line validation}

\begin{figure}[tb]
\centering
  a)\,\includegraphics[width=0.44\linewidth]{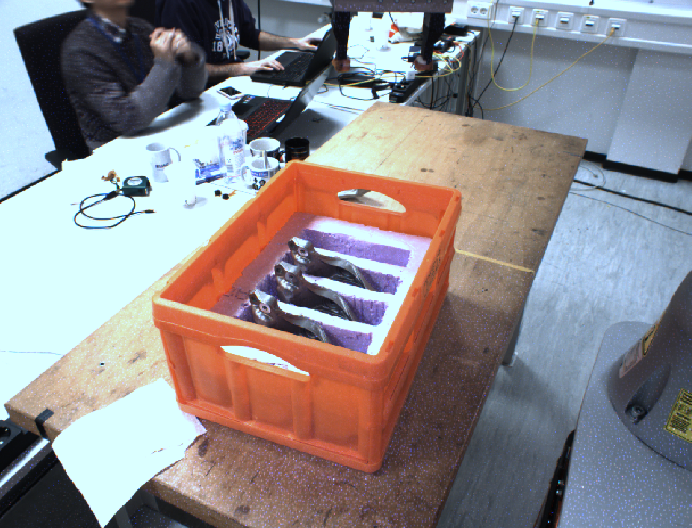}\hspace*{3ex}
  b)\,\includegraphics[width=0.44\linewidth]{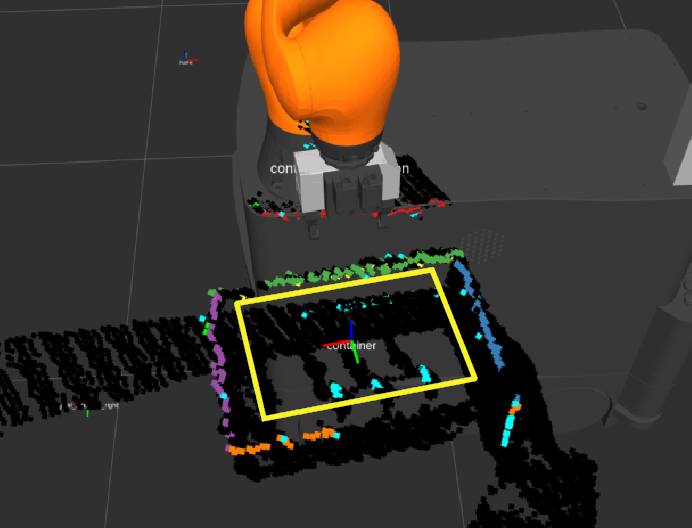} \vspace*{-1ex}
  \caption{Box detection.
  a) Raw image from the pan-tilt camera;
  b) Point cloud with detected edge points (cyan) and line segments (random colors). Best matched model is shown as yellow rectangle.}
  \label{fig:lines}
\end{figure}

After the line detection, we perform a validation step which is based on two restrictions:
\begin{itemize}
	\item {\em Connectivity Restriction.}
	The inliers of a detected line may lie on different unconnected line segments.
	While partial occlusions can cause multiple unconnected line segments on the edges of the box, 
	we cluster the inliers and split the detected line into multiple segments in case the box should be fully visible.
	If the number of points in a cluster falls below the minimum number of inliers for line segments, it is neglected.
	\item {\em Length Restriction.}
	Line segments which are shorter than the shortest edge in the model and longer than the longest edge in the model are neglected.
	To account for noise, missing edge points, or other errors, a deviation of 20\% from these thresholds is allowed.
\end{itemize}
An example of detected edges and lines is shown in \reffig{fig:lines}.

\subsubsection{Model sampling and validation}

In order to detect the container, a subset of the detected line segments is selected.
We select $N$ line segments where $N$ is the number of line segments in the parametrized model.
That is, we sample as many line segments as contained in the model of the container.
As a result, we obtain tuples of line segments $(l_0, ..., l_N)$.
To avoid repetitively re-checking the same tuples, we use a hash table in which sampled tuples are marked as being processed.

We discard tuples of line segments which are not compatible with the container model.
The model contains four edges which are pairwise parallel and perpendicular to each other.
If the tuple of sampled line segments is valid, we continue to register the model against the sampled line segments.
For the model registration, we sample points from the given parametrized container model in order to obtain a source point cloud $P$ for registration.
In addition, we extract the inliers of the sampled segments to form a single target point cloud $Q$ for registration.
In contrast to extracting the inliers for the target point cloud, the source point cloud of the model only needs to be sampled  once.

Iterative registration algorithms align pairs of 3D point clouds by alternately searching for correspondences between the clouds and minimizing the distances between matches~\cite{RAM:Holz2015}.
In order to align a point cloud $P$ with a point cloud $Q$, the ICP algorithm searches for the closest neighbors in $Q$ for points $p_i \in P$ and minimizes the point-to-point distances
$d^T_{ij} = q_j - Tp_i$ of the set of found correspondences $C$  in order to find the optimal transformation $T^*$:
\begin{equation}\label{eq:transformation}
  T^* = \argmin_{T} \sum_{(ij) \in X} {|| d^{(T)}_{ij} ||}^2.
\end{equation}
Finally, we compute a confidence $c$ that is based on the overlap between the model and the sampled line segments: $c=|C|/|P|$,
where $|C|$ is the number of corresponding points within a predefined distance tolerance $\epsilon_d$, and $|P|$ is the number of points in the generated model point cloud. 
In case of a complete overlap, the confidence $c$ is roughly 1.
We select the best match to estimate the pose of the container.

\subsection{Part Segmentation}
\label{sec:parts_segmentation}

In order to segment the parts that come in the boxes, we use the estimated pose of the box from the previous step.
This gives us an observation pose for the wrist camera above the box, which is used to have a clear view of the inside of the container.
We use the detected box borders to extract the points in the obtained point cloud that correspond to the contents of the box.

\begin{figure}[tb]
\centering
  a)\,\includegraphics[height=0.272\linewidth]{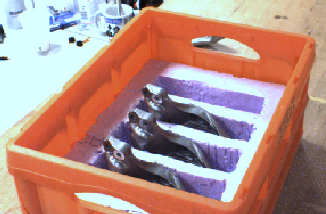}\hspace*{3ex}
  b)\,\includegraphics[height=0.272\linewidth]{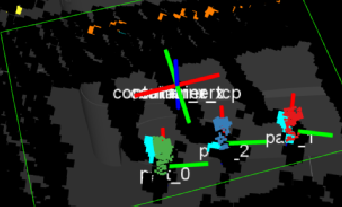}\vspace*{-2ex}
  \caption{Part segmentation.
  a) Raw image from the pan-tilt camera. 
  b) Three segmented parts (green, blue and red) with their corresponding grasping poses.}
  \label{fig:parts_segmentation}
\end{figure}

Engine support variants are segmented using Euclidean clustering on the box content point cloud.
The centroid and principal axes of the clusters are used to compute the grasping poses.
An example of the segmented parts is shown in \reffig{fig:parts_segmentation}

To determine grasps for the engine pipes in the extracted container content, we cluster the extracted point cloud into cylinders 
and select the centroid of the highest cylinder as the grasping point.
The orientation of the grasp pose is chosen according to the principal axis of the cylinder to be grasped and aligned with the local coordinate frame
of the detected container in order to approach the part straight from the top of the container.

\subsection{Parts Variant Recognition}
\label{sec:part_recognition}

In real production warehouses, the locations of containers with parts may be mixed up, or a part of a wrong type may accidentally enter a container with other parts.
In order to detect such situations, we perform part recognition.
The recognition takes place after the part was grasped and lifted up in the air, since in such position it is unoccluded and may be easily observed by the PTSS.

\begin{figure}[t]
  \centering
     \includegraphics[width=\linewidth]{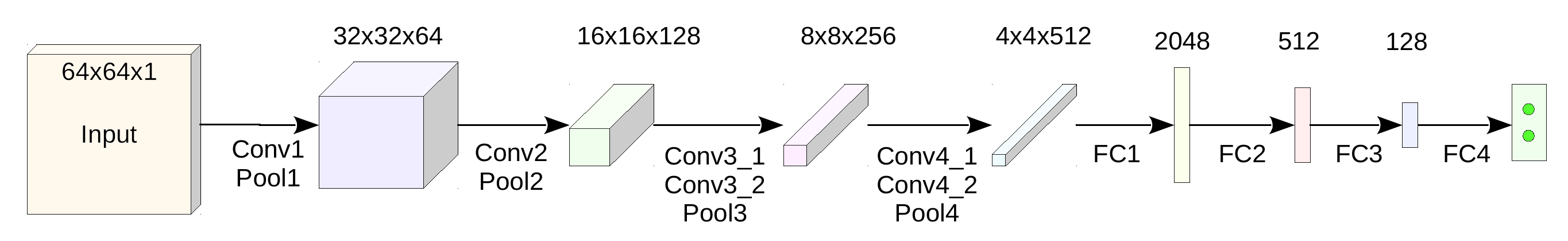}\vspace*{-2ex}
  \caption{Neural network used to recognize the variant of an engine support part.}
  \label{fig:nn}
\end{figure}

A convolutional neural network, shown in \reffig{fig:nn}, is used to perform the recognition.
The network takes a 64$\times$64$\times$1 depth image as input.
Metal parts are shiny and thus shape features may be not visible on the RGB image.
The use of depth information helps to overcome this issue.
The first part of the network consists of four convolutional layers, each followed by a pooling layer.
The final part consists of four fully connected layers.
The last layer outputs two values through a softmax function.
The numbers represent the probability of the object belonging to the first and the second class, respectively.
We used this network to distinguish between two types of engine supports, as these parts look similar and could be mixed.

In order to train the model, available CAD models of the parts were used to render depth maps.
For each variant, 10,000 different poses were used in order to produce synthetic data.
To obtain an input to the network in the real world, we project the center of the TCP to the pan-tilt depth image and crop an image window that contains the part.

\section{Manipulation}
\label{sec:manipulation}

Given a grasp pose from the perception module, it is necessary to plan a trajectory for the robotic arm to reach the corresponding pre-grasp pose.
The trajectory has to be smooth and must avoid any collisions with the environment or the robot itself.
Furthermore, it has to satisfy constraints on orientation of the end-effector.
Moreover, the duration of the trajectory has to be as short as possible, since it directly influences the overall time spent for the kit completion.
For the same reason, planning time must be short.
Finally, it is necessary to constantly track the future part of the trajectory during execution to detect any unforeseen collisions with dynamic objects.
In case when future collision is detected, the trajectory has to be replanned as fast as possible and the execution should continue.

To fulfill these requirements, we use our trajectory optimization method~\cite{Pavlichenko2017} which is based on STOMP~\cite{Kalakrishnan2011}.
The method iteratively samples noisy trajectories around a mean trajectory and evaluates each of them with a cost function.
Then the mean is shifted in the direction of reducing costs.
Iterations continue until one of the termination criteria is met.

The trajectory $\Theta$ is defined as a sequence of $N$ keyframes in joint space.
Start and goal configurations are fixed, as well as the number of keyframes $N$.
The cost of the trajectory $\Theta$ is defined as a sum of costs of transitions between adjacent keyframes $\bm{\theta}_{i}$:\vspace*{-1ex}
\begin{equation}\label{eq:traj_opt:cost}
q(\Theta) = \sum_{i=0}^{N-1} q(\bm{\theta}_i, \bm{\theta}_{i+1}).
\end{equation}
We propose a cost function which consists out of five components:
\begin{equation}\label{eq:cost_function}
\begin{aligned}
q(\bm{\theta}_{i}, \bm{\theta}_{i+1}) =  &q_{o}(\bm{\theta}_{i}, \bm{\theta}_{i+1}) + q_l(\bm{\theta}_{i}, \bm{\theta}_{i+1}) + q_{c}(\bm{\theta}_{i}, \bm{\theta}_{i+1}) \\ + &q_{d}(\bm{\theta}_{i}, \bm{\theta}_{i+1}) + q_{t}(\bm{\theta}_{i}, \bm{\theta}_{i+1}),
\end{aligned}
\end{equation}
where 
$q_o(\bm{\theta}_{i}, \bm{\theta}_{i+1})$ is a component which penalizes being close to obstacles,
$q_l(\bm{\theta}_{i}, \bm{\theta}_{i+1})$ penalizes exceeding of joint limits,
$q_c(\bm{\theta}_{i}, \bm{\theta}_{i+1})$ penalizes task specific constraints on a gripper position or/and orientation,
$q_d(\bm{\theta}_{i}, \bm{\theta}_{i+1})$ penalizes long durations of the transitions between the keyframes and
$q_t(\bm{\theta}_{i}, \bm{\theta}_{i+1})$ is a component that penalizes high actuator torques.
Each cost component $q_j(.,.)$ is normalized to be within $[0,1]$ interval and has an importance weight $\lambda_j \in [0,1]$ attached.
This allows to prioritize optimization by manipulating weights $\lambda_j$.

In order to speed up the optimization process, we utilize two phases.
During the first phase, a simplified cost function is used. It consists of collision costs $q_o$, joint limit costs $q_l$, and gripper constraints costs $q_c$.
As soon as the first valid solution is found, the second phase begins, where the full cost function $q(.,.)$ (as described in Equation~\ref{eq:cost_function}) is used.
Optimization continues until one of the termination criteria is met.

\section{Experiments}
\label{sec:experiments}

In order to assess the designed system, we performed several experiments during the EuRoC Showcase evaluation in the lab of the Challenge~2 host DLR Institute of Robotics and Mechatronics in Oberpfaffenhofen, Germany, under severe time constraints and the supervision of judges.
The experiments included tests of isolated components as well as full kitting procedures.
All experiments have been done on the real robot.
In this section, we describe the test procedures and present obtained results.

\begin{figure}[t]
  \centering
     \includegraphics[width=10cm]{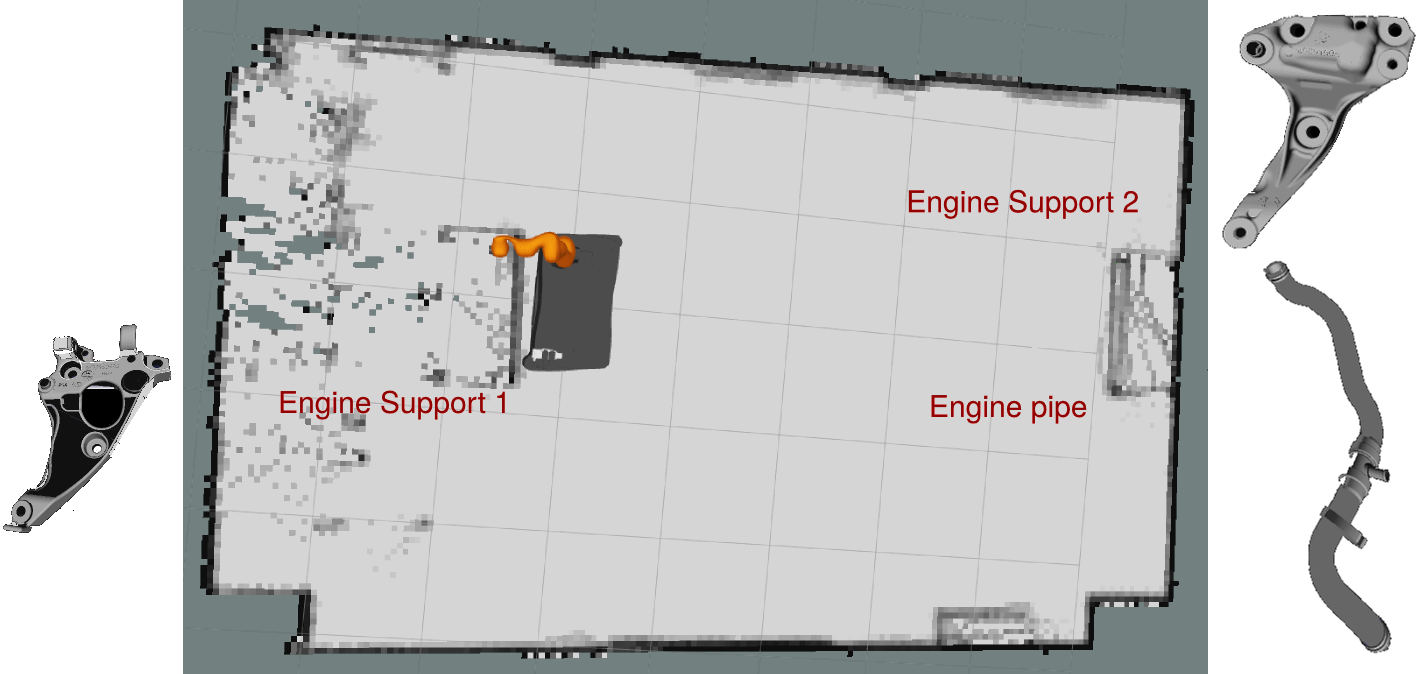}\vspace*{-1ex}
  \caption{Map of the automotive supermarket. A container with Engine Support~1 parts is located on the first table.
	   Containers with Engine Support~2 parts and Engine Pipes are located on the second table. CAD models of the parts are shown on the sides of the map.}
  \label{fig:supermarket}
\end{figure}

\subsection{Showcase Setup}
\label{sec:setup}

The kitting experiment was performed in a simplified supermarket of parts, designed by us.
The kit consisted out of three automotive parts, supplied by our end-user partner PSA:
\begin{itemize}
	\item Engine Support~1: metal part with shiny surface.
	\item Engine Support~2: metal part with shiny surface, very similar to Engine Support~1.
	\item Engine Pipe: black flexible pipe made out of rubber.
\end{itemize}
Each part type is provided in a separate container.
Both engine supports were placed in the containers with slots, so that each part is positioned roughly vertically, perpendicular to the bottom of the container.
Engine pipes were put into their container without any order, making picking more challenging.

The map of the automotive supermarket as well as CAD models of the parts are show in \reffig{fig:supermarket}.
Containers with parts were located on tables in opposite sides of the 10$\times$5\,m room.
The container with Engine Supports~1 was located on the first table.
Containers with Engine supports~2 and Engine pipes were provided on the second table.

\subsection{Kitting}
\label{sec:kitting}

The procedure of our kitting scenario was defined as follows: the robot starts in the middle of the supermarket.
It has to move to the first table and pick up Engine Support~1 and place it in the first kitting compartment on the robot.
After that, the robot has to move to the second table and pick up Engine Support~2 and place it in the second kitting compartment.
Finally, the robot has to pick up Engine Pipe and place it in the third kitting compartment.
To demonstrate that the kit is ready to be delivered to the assembly line, the robot moves away from the table.

\begin{figure}[t]
\centering \footnotesize
  a)\,\includegraphics[height=0.215\linewidth]{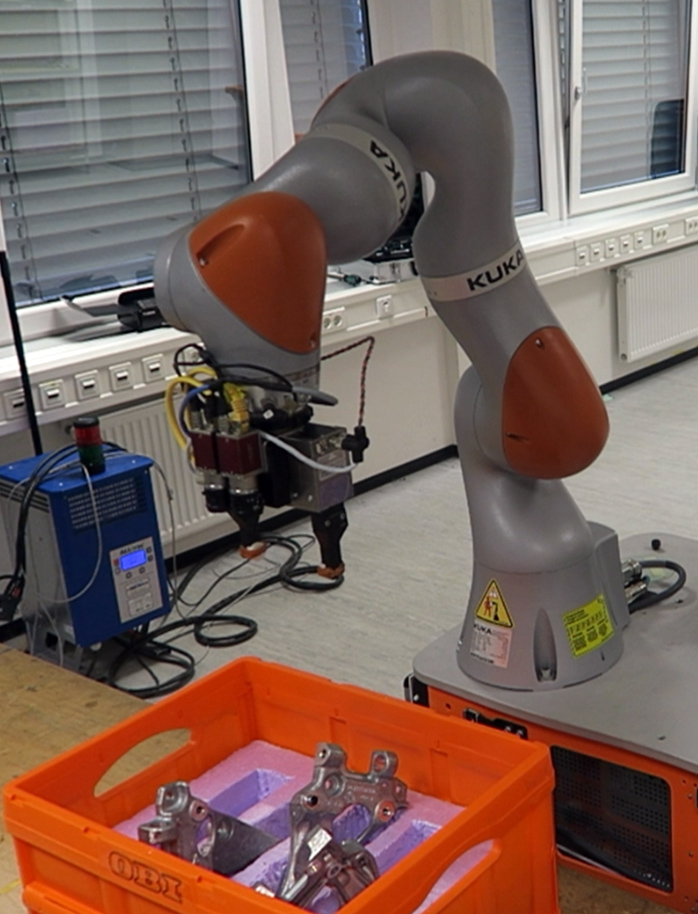}  b)\,\includegraphics[height=0.215\linewidth]{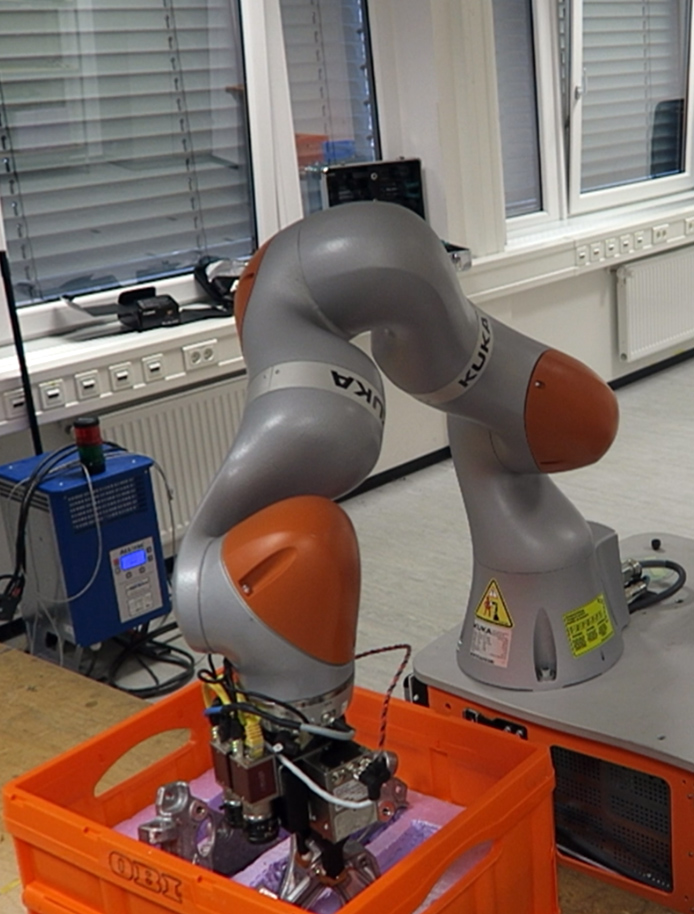}  c)\,\includegraphics[height=0.215\linewidth]{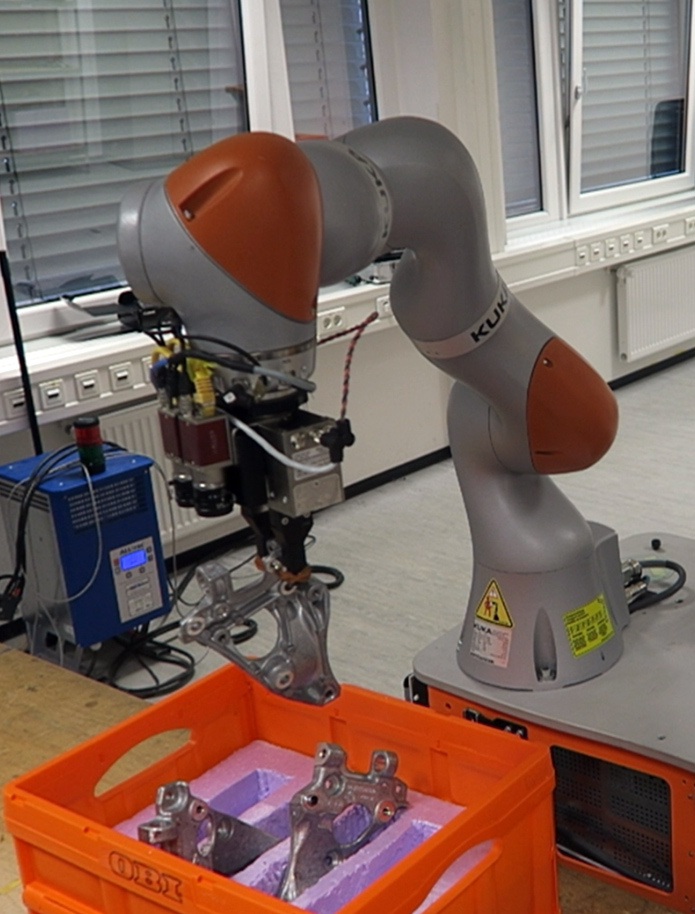}  d)\,\includegraphics[height=0.215\linewidth]{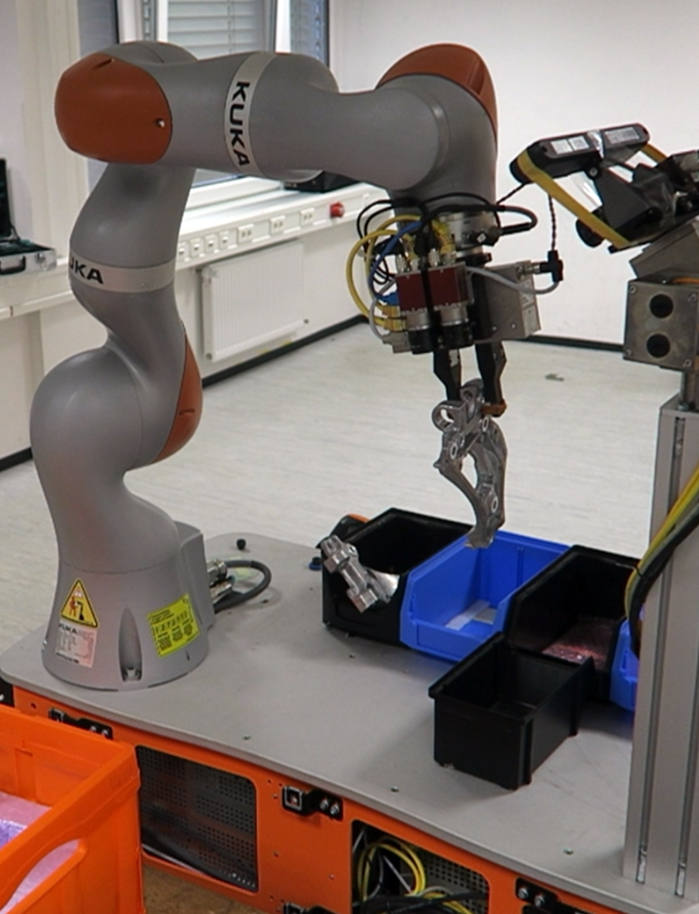}  e)\,\includegraphics[height=0.215\linewidth]{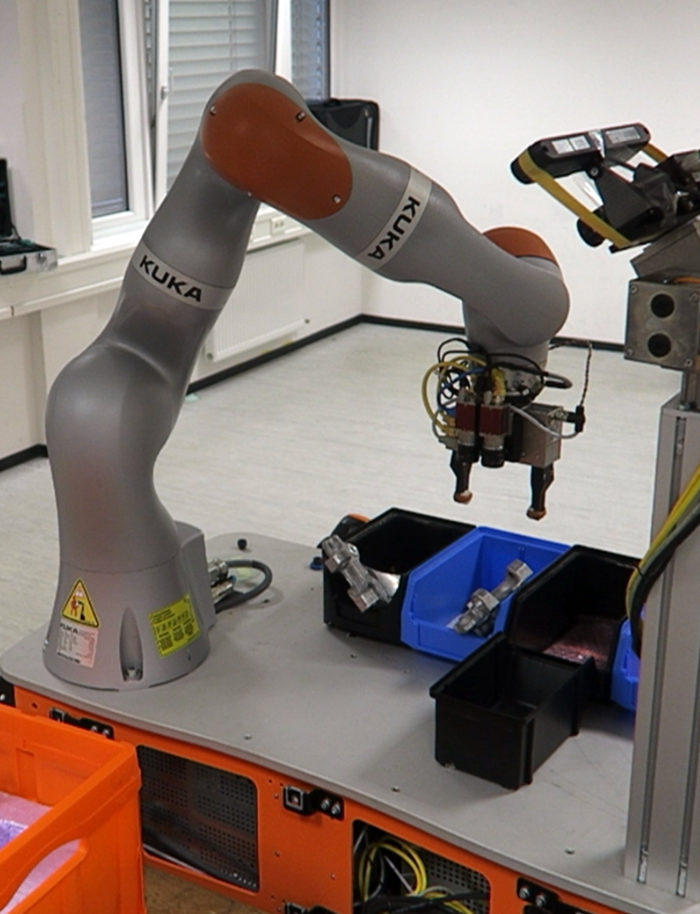}
  \caption{Picking of Engine Support~1. 
  a) Observation pose; 
  b) Part grasped; 
  c) Part lifted;
  d) Part transported to the drop pose; 
  e) Part placed into the kitting compartment on the robot.}
  \label{fig:picking}
\end{figure}

In order to demonstrate the capability of our system, we performed two kitting runs, as described above.
The robot picking the Engine Support~1 is shown in \reffig{fig:picking}.
Videos of the experiments are available online\footnote{Experiment video: \url{http://www.ais.uni-bonn.de/videos/IAS_2018_KittingBot}}.
We measured the success rate of picking and placing for each part type, as well as the overall runtime.
The results are presented in the Table~\ref{table:results}.
One can observe that placing the parts never failed.
Picking parts succeeded on all but one case: picking of Engine Pipe in the first run was not successful.
The placing task is much easier than picking, since the positions of the kitting boxes on the robot are known precisely.
Picking of the engine pipe failed because the robot attempted to grasp it above the widest part of the pipe.
Consequently, the grasp was not firm enough and the part slipped from the gripper.

\begin{table}[tbh]
\centering \vspace*{-2ex}
\caption{Pick and place success rates and runtime of kitting.}
\label{table:results}
\vspace*{1ex}\begin{tabular}{lcc}
\hline
                                   & \textbf{     Run 1     } & \textbf{     Run 2     } \\ \hline
\textbf{Parts successfully picked} & 2/3            & 3/3            \\ \hline
Engine Support~1                   & +              & +              \\ \hline
Engine Support~2                   & +              & +              \\ \hline
Engine Pipe                        & -              & +              \\ \hline
\textbf{Successful grasps}         & 2/3            & 3/3            \\ \hline
\textbf{Successful placements}     & 2/2            & 3/3            \\ \hline
\textbf{Runtime {[}s{]}}           & 759            & 809            \\ \hline
\end{tabular}\vspace*{-2ex}
\end{table}

\subsection{Additional Experiments}
\label{sec:additional_expiriments}

In addition to the complete kitting procedure, we tested several components in isolation.
In this subsection we present the obtained results.

\subsubsection{Part Variant Recognition.}
\begin{figure}[tb]
\centering
  a)\,\includegraphics[width=0.44\linewidth]{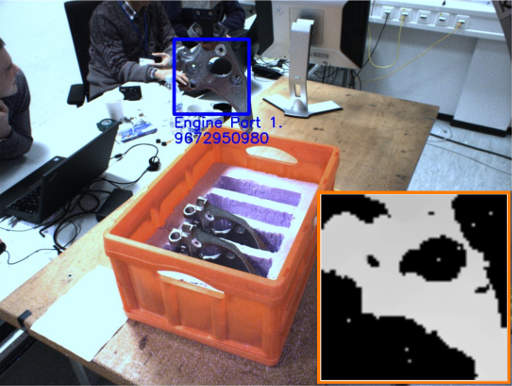}\hspace*{3ex}
  b)\,\includegraphics[width=0.44\linewidth]{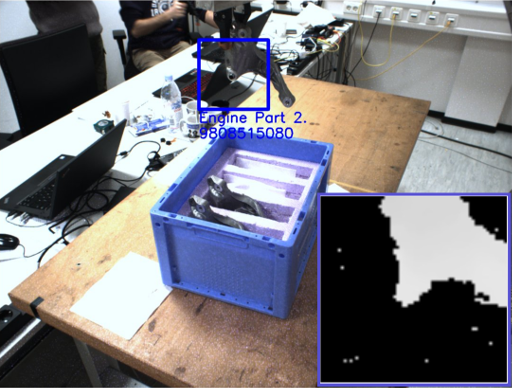} \vspace*{-1ex}
  \caption{Recognition of the part variant. 
  a) Raw image of the picked up Engine Support~1; input depth image shown in the bottom-right corner. 
  b) Raw image of the picked up Engine Support~2; input depth image in the bottom-right corner.}
  \label{fig:recognition}
\end{figure}

In this experiment, we demonstrated the capabilities of part variant recognition module.
First, we pick up Engine Support~1 and recognize which part is in the gripper.
Then we pick up Engine Support~2 and perform the recognition again.
The obtained images of the parts are shown in \reffig{fig:recognition}.
Both parts were recognized correctly.

\subsubsection{Unforeseen Collision Avoidance.}
\begin{figure}[t]
\centering
  a)\,\includegraphics[width=0.4\linewidth]{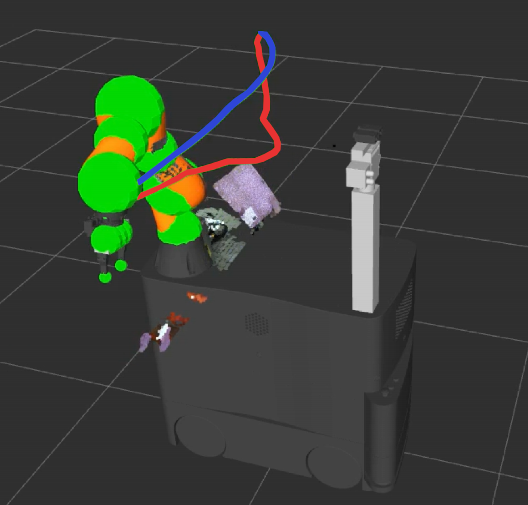}\hspace*{3ex}
  b)\,\includegraphics[width=0.4\linewidth]{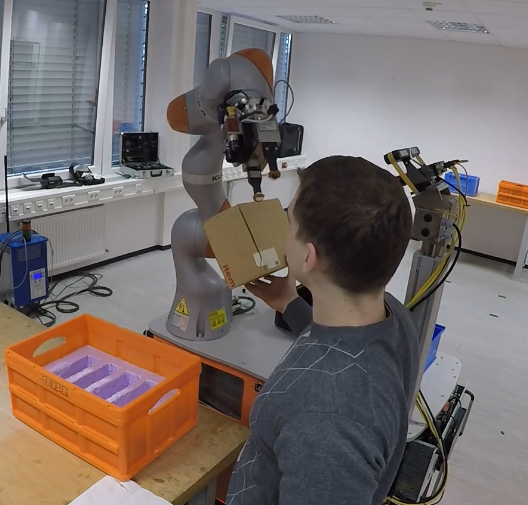}\vspace*{-1ex}
  \caption{
  Replanning of the trajectory to avoid an unforeseen obstacle. 
  a) The obstacle is inserted during execution. The trajectory is replanned. Red: initial trajectory. Blue: replanned trajectory. 
  b) Arm, avoiding the new obstacle.}
  \label{fig:replanning}
\end{figure}

To demonstrate the ability of our system to deal with obstacles which appear during trajectory execution, we performed a separate experiment.
The robot arm had to move from the observation pose to the pose above the kitting boxes.
After the trajectory had been planned and the execution was started, an obstacle was inserted on the way.
The system continuously tracks the future part of the trajectory and checks for obstacles during trajectory execution.
The future collision was detected, the execution was stopped and the trajectory was replanned, taking the new obstacle into consideration.
Both initial and replanned trajectories, as well as the arm avoiding the obstacle are shown in \reffig{fig:replanning}.
The replanning took 0.39 seconds, which in principle allows to perform replanning without stopping the execution,
in case the arm does not move too fast and the collision is far enough ahead.

\section{Conclusion}
\label{sec:conclusion}

We have developed a mobile manipulation system system for performing autonomous part kitting.
We proposed perception software modules which allow to efficiently detect containers, segment the parts therein, and produce grasps.
In addition, our system is capable of recognizing part variants.
The developed manipulation planner is able to optimize robotic arm trajectories with respect to collisions, joint limits, end-effector constraints, joint torque, and duration.
The method allows to perform optimization fast and to replan trajectories in case of possible future collisions due to newly appeared obstacles.
We integrated these modules into a Kuka KMR iiwa robot.
Together with the Kuka navigation stack, our components formed a system capable of autonomous kitting under guidance of a high-level FSM.

We demonstrated the capabilities of our system in a simplified kitting scenario during the EuRoC Showcase evaluation, in the lab of the challenge host, supervised by judges under severe time constraints. 
The experiments shown that the perception module can reliably detect containers and segment the parts inside them.
Generated gasps were reliable in the most cases, failing only once when grasping an Engine Pipe, which was the hardest part in the kit.
The trajectory optimization method shown good performance with short runtimes and allowed to deliver the parts to the kitting compartments in all cases.
Real-time supervision of the workspace and online replanning are a suitable basis for collaborative kitting.

\vspace*{1ex}
{\footnotesize\noindent\textbf{Acknowledgements.} This research received funding from the European Union's Seventh Framework Programme  grant agreement no. 608849 (EuRoC). 
It was performed in collaboration with our end-user partner Peugeot Citro\"{e}n Automobiles S.A. (PSA).
We also gratefully acknowledge the support of the EuRoC Challenge~2 host: DLR Institute of Robotics and Mechatronics in Oberpfaffenhofen, Germany.}

\bibliography{references}{}
\bibliographystyle{splncs}

\end{document}